\documentclass{article}
\usepackage{ijcai22}

\usepackage{times}
\usepackage{soul}
\usepackage{url}
\usepackage[hidelinks]{hyperref}
\usepackage[utf8]{inputenc}
\usepackage[small]{caption}
\usepackage{graphicx}
\usepackage{amsmath}
\usepackage{amsthm}
\usepackage{amsfonts,bm}
\usepackage{booktabs}
\usepackage{algorithm}
\usepackage{algorithmic}
\usepackage{subfigure}
\usepackage{enumitem}
\usepackage{multirow}
\newtheorem{mydef}{Definition}
\newtheorem{mytheo}{Theorem}

\newenvironment{myproof}{{\noindent\it Proof.}\noindent}{\hfill $\square$\par}

\title{A Simple yet Effective Method for Graph Classification}

\author{
Junran Wu$^{1\ast}$\and 
Shangzhe Li$^{1,2}$\footnote{Equal Contribuation.}\and
Jianhao Li$^{1}$\and 
Yicheng Pan$^{1\dag}$\And 
Ke Xu$^{1}$\footnote{Correspondence to: Yicheng Pan, Ke Xu.}\\
\affiliations
$^1$State Key Lab of Software Development Environment, Beihang University, Beijing, 100191, China\\
$^2$School of Mathematical Science, Beihang University, Beijing 100191, China\\
\emails
\{wu\_junran, shangzheli, lijianhao, yichengp, kexu\}@buaa.edu.cn}

\begin{document}

\maketitle

\begin{abstract}
In deep neural networks, better results can often be obtained by increasing the complexity of previously developed basic models. However, it is unclear whether there is a way to boost performance by decreasing the complexity of such models. 
Intuitively, given a problem, a simpler data structure comes with a simpler algorithm.
Here, we investigate the feasibility of improving graph classification performance while simplifying the learning process.
Inspired by structural entropy on graphs, we transform the data sample from graphs to coding trees, which is a simpler but essential structure for graph data.
Furthermore, we propose a novel message passing scheme, termed hierarchical reporting, in which features are transferred from leaf nodes to root nodes by following the hierarchical structure of coding trees.
We then present a tree kernel and a convolutional network to implement our scheme for graph classification.
With the designed message passing scheme, the tree kernel and convolutional network have a lower runtime complexity of $O(n)$ than Weisfeiler-Lehman subtree kernel and other graph neural networks of at least $O(hm)$. We empirically validate our methods with several graph classification benchmarks and demonstrate that they achieve better performance and lower computational consumption than competing approaches.
\end{abstract}

\section{Introduction}
\label{sec:intro}
Over the years, deep learning has achieved great success in perception tasks, such as recognizing objects or understanding language, which are difficult for traditional machine learning methods \cite{bengio2021deep}. To further enhance performance, research efforts have generally been devoted to designing more complex models based on previously developed basic ones; such improvements include increasing model depths (e.g., ResNet \cite{he2016deep}), integrating more complicated components (e.g., Transformer \cite{vaswani2017attention}) or even both (e.g., GPT3 \cite{brown2020language} and NAS \cite{liu2020block}). However, little work has focused on the research direction of boosting performance through simplifying the basic model learning process.

\begin{figure*}[!t]
  \centering
  \includegraphics[width=1\textwidth]{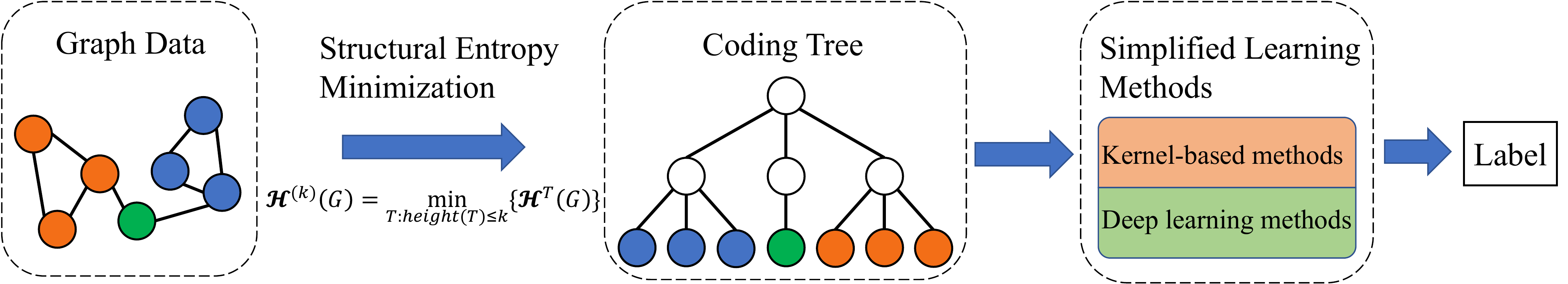}   
  \caption{\textbf{Overview of graph simplification via structural entropy minimization.} Structural entropy guides the simplification for graphs to decode the underlying structure, and several learning algorithms can benefit from it.}
  \label{fig:semp_overview}
\end{figure*}

Similarly, there are many interesting tasks involving graphs that are hard to learn with normal deep learning models, which prefer data with a grid-like structure. Thus, graph neural networks (GNNs) have recently been ubiquitous within deep learning for graphs, and achieve great success in various domains because of their ability to model structural information \cite{hamilton2017inductive,zhang2018end,xu2019powerful}.
In order to pursuit more superior performance, various more complex components have also been developed and integrated into basic GNNs \cite{velivckovic2018graph,ying2018hierarchical,zhang2018end}.
In this context, the improvement in performance comes at the price of model complexity, analogous to routines in deep learning.
Here, we investigate the feasibility of improving graph classification performance by simplifying model learning. Generally, given a problem, a simpler data structure comes with a simpler algorithm.
Motivated by structural entropy \cite{li2016structural}, a metric designed to assess the structural information of a graph, the essential structure of a graph can be decoded by this metric as a measure of the complexity of its hierarchical structure. 
Therefore, as shown in Fig. \ref{fig:semp_overview}, we propose an algorithm to simplify the given data sample of a graph by transforming it into a corresponding coding tree that reflects the hierarchical organization of data, in which the crucial structural information underlying the graph can be retained in the coding tree with minimal structural entropy.

Based on the simplified coding trees, we propose a novel feature combination scheme for graph classification, termed hierarchical reporting. In this scheme, we transfer features from leaf nodes to root nodes based on the hierarchical structures of the associated coding trees. For two kinds of basic learning algorithms (i.e., kernel-based methods and GNNs), we propose two corresponding simplified learning algorithms; put differently, we present an implementation of our scheme in a tree kernel and a convolutional network, denoted the Weisfeiler-Lehman coding tree (WL-CT) kernel and hierarchical reporting network (HRN), to perform graph classification. The tree kernel follows the label propagation in the Weisfeiler-Lehman (WL) subtree kernel but has a lower runtime complexity of $O(n)$. HRN is an implementation of our tree kernel in the deep learning field. Finally, we empirically validate our WL-CT kernel and HRN on various graph classification datasets. Our tree kernel surpasses the state-of-the-art kernel-based methods and even outperforms GNNs on several benchmarks. Our HRN also achieves state-of-the-art performance on most benchmarks. 

We list our contributions in this work as follows:
\begin{itemize}
    \item We present a novel direction to boost the performance of learning models while reducing complexity.
    \item With structural entropy minimization, we optimize the given data sample from graphs to coding trees, which are much simpler data structures that also retain the crucial structural information of graphs.
    \item We develop two efficient learning algorithms, i.e., WL-CT and HRN, and empirically present their discriminative power on many graph classification benchmarks.
\end{itemize}

\section{Related Work}
\label{sec:liter}
GNNs have achieved state-of-the-art results on various tasks with graphs, such as node classification \cite{velivckovic2018graph}, link prediction \cite{zhang2018link} and graph classification \cite{hamilton2017inductive,zhang2018end,xu2019powerful}. In this work, we devote our attention to graph classification scenarios.

Graph classification involves identifying the characteristics of an entire graph and is ubiquitous in a variety of domains, such as social network analysis \cite{backstrom2011supervised}, chemoinformatics \cite{duvenaud2015convolutional}, and bioinformatics \cite{borgwardt2005protein}.
In addition to previous techniques such as graph kernels \cite{shervashidze2011weisfeiler}, GNNs have recently emerged and become a popular way to handle graph-related tasks due to their effective and automatic extraction of graph structural information \cite{hamilton2017inductive,zhang2018end,xu2019powerful}.
To address the limitations of various GNN architectures, the graph isomorphism network (GIN) \cite{xu2019powerful} was presented in theoretical analyses regarding the expressive power of GNNs in terms of capturing graph structures.
All GNNs are broadly based on a recursive message passing (or neighborhood aggregation) scheme, where each node recursively updates its feature vector with the ``message'' propagated from neighbors \cite{gilmer2017neural,xu2019powerful}. The feature vector representing an entire graph for graph classification can be obtained by a graph pooling scheme \cite{ying2018hierarchical}, such as the summation of all node feature vectors of the graph. Accordingly, much effort has been devoted to exploiting graph pooling schemes, which are applied before the final classification step \cite{zhang2018end,ying2018hierarchical}.
All these pooling methods help models achieve state-of-the-art results but increase the model complexity and the volume of computations.

Structural entropy is a measure of the complexity of the hierarchical structure of a graph \cite{li2016structural}. The structural entropy of a graph is defined as the average length of the codewords obtained under a specific coding scheme for a random walk. That is, when a random walk takes one step from node $u$ to node $v$, the codeword of the longest common ancestor of $u$ and $v$ on the coding tree, which is also their longest common prefix, is omitted. This shortens the average codeword length. Equivalently, the uncertainty of a random walk is characterized by this value, which is the origin of the term structural entropy. The coding tree of a graph that achieves the minimum structural entropy indicates the optimal hierarchical structure. Furthermore, two- and three-dimensional structural entropy, which measure the complexity of two- and three-level hierarchical structures, respectively, have been applied in bioinformatics \cite{li2018decoding}, medicine \cite{li2016three}, the structural robustness and security of networks \cite{li2016resistance}, etc. For tasks regarding graphs, structural entropy can be used to decode the essential structure as a measure of the complexity of its hierarchical structure.

Thus far, there has been little attention paid to the model promotion and efficiency rising through data sample simplification. In this paper, with the guidance of structural entropy, we investigate the feasibility of improving graph classification performance with lower model complexity.

\section{Methodology}
\label{sec:sturct_opt}
In this section, we first introduce an algorithm for graph simplification in which we decode the essential structure of an input graph to the corresponding coding tree by minimizing its structural entropy. 
Based on the coding trees, we propose a tree kernel and a corresponding implementation in a deep learning model for graph classification. We elaborate on them in the following.

\subsection{Graph Simplification via Structural Entropy Minimization}
According to the definition in \cite{li2016structural}, given a graph $G=(V, E)$ and a coding tree $T$ for $G$, where $|V|=n$ and $|E|=m$, the structural entropy of $G$ on $T$ is defined as
\begin{equation}
\mathcal{H}^T (G)=-\sum_{v_t\in T} \frac{g_{v_t}}{vol(V)} \log \frac{vol(v_t)}{vol(v_t^+)},
\end{equation}
where $v_t$ is a nonroot node in $T$ that also represents a node subset $\subset V$, $v_t^+$ refers to the parent node of $v_t$, $g_{v_t}$ is the number of edges with exactly one endpoint in $v_t$, and $vol(V)$ and $vol(v_t)$ are the sums of degrees of nodes in $V$ and $v_t$, respectively. 
The structural entropy of $G$ is defined to be the minimum entropy among all coding trees, and it is denoted by $\mathcal{H}(G)=\min_{T}\{\mathcal{H}^T (G)\}.$ 

To formulate a natural coding tree with a certain height, the $k$-dimensional structural entropy of $G$ for any positive integer $k$ is the minimum value among all coding trees with heights of at most $k$:
\begin{equation}
\mathcal{H}^{(k)}(G)=\min_{T:\text{height}(T)\leq k}\{\mathcal{H}^T (G)\}.
\end{equation}
In this paper, to realize the simplification of given graphs with a certain $k$-dimensional structural entropy, we propose a greedy algorithm to compute the $k$-dimensional coding tree with minimum structural entropy in Algorithm \ref{code:coding_tree}.
Specifically, we first build a full-height binary coding tree from an initial coding tree. In this phase, we iteratively merge two nonroot nodes, which can minimize the structural entropy at each step. Then, to form the $k$-dimensional coding tree $T$ for $G$, we repeatedly delete one edge from $T$, which can minimize the structural entropy restoration at each step. Finally, a coding tree $T$ with a certain height $k$ for $G$ can be obtained, where $T=(V_T, E_T)$, $V_T=(V_T^0, \dots, V_T^k)$ and $V_T^0=V$. 

The time complexity of $k$-dimensional coding tree is $O(h_{max}(m\log n+n))$, where $h_{max}$ is the maximum height of coding tree $T$ during the generation of the full-height binary coding tree. Since the minimized structural entropy tends to construct balanced coding trees, $h_{max}$ is generally of order $O(\log n)$. Furthermore, the number of edges of graph is generally greater than the number of nodes, thus the runtime of Algorithm \ref{code:coding_tree} almost scales linearly in the number of edges.

\begin{algorithm}[!t]
\caption{$k$-dimensional coding tree on structural entropy}
\label{code:coding_tree} 
\textbf{Input:} a graph $G=(V, E)$, a positive integer $k>1$\\
\textbf{Output:} a $k$-dimensional coding tree $T$

\begin{algorithmic}[1]
\STATE Initialize $T$ with one root node $v_r$ and all nodes in $V$ as leaves;
\STATE Define the function $merge(v_i, v_j)$ for $T$ to insert a new node between $v_r$ and $(v_i, v_j)$;
\STATE Define the function $delete(v_p, v_c)$ for $T$ to delete $v_c$ from $T$ and link $v_c.children$ to $v_p$, where $v_c.parent=v_p$;
\STATE // Generate the full-height binary coding tree $T$ for $G$;
\WHILE{$|v_r.children|>2$} {
  \STATE Select $v_i$ and $v_j$ from $v_r.children$, which make \\
  $argmax_{(v_i, v_j)}\{\mathcal{H}^T(G) - \mathcal{H}^{T_{merge(v_i, v_j)}}(G)\}$;
  \STATE Execute $merge(v_i, v_j)$;
}
\ENDWHILE
\IF{Height($T$) $\leq k$} {
  \STATE return $T$;
}
\ENDIF
\STATE // Compress $T$ to height $k$;
\WHILE{Height($T$) $>k$} {
  \STATE Select $v_p$ and $v_c$ from $T$, which make \\
  $argmin_{(v_p, v_c)}\{\mathcal{H}^{T_{delete(v_p, v_c)}}(G) - \mathcal{H}^T(G)\}$;
  \STATE Execute $delete(v_p, v_c)$;
}
\ENDWHILE
\STATE return $T$;
\end{algorithmic} 
\end{algorithm}

\subsection{A Tree Kernel for Graph Classification}
To perform classification on coding trees, following the construction of the WL subtree kernel \cite{shervashidze2011weisfeiler}, we propose a novel tree kernel that measures the similarity between coding trees named the WL-CT kernel. The key difference between the two kernels is the label propagation scheme, where we develop a \textit{hierarchical reporting} scheme to propagate labels from child nodes to their parents based on the hierarchical structures of coding trees. Finally, our tree kernel adopts the counts of node labels of the entire coding tree as the feature vector of the original graph.

\paragraph{Hierarchical reporting.}The key idea of this scheme is to assign labels to nonleaf nodes by aggregating and sorting the labels from their child nodes and then to compress these sorted label sets into new and short labels. Labels from the leaf nodes are iteratively propagated to the root node, which means that the iteration time of this scheme is determined by the height of the coding tree, and naturally avoids the convergence problem of WL subtree kernel.

\begin{mydef}
\label{def:tree_kernel}
Let $T_1$ and $T_2$ be any two coding trees with the same height $k$. There exists a set of letters $\Sigma^i \in \Sigma$, which are node labels appearing at the $i$-th ($i<k$) layer of $T_1$ or $T_2$ (i.e., the nodes with height $i$ are assigned labels with hierarchical reporting). $\Sigma^0$ is the set of leaf node labels of $T_1$ and $T_2$. Assume that any two $\Sigma^i$ are disjoint, and every $\Sigma^i = \{b^i_1,\dots,b^i_{\left|\Sigma^i\right|}\}$ is ordered without a loss of generality. We define a function $c^i: \{T_1,T_2\}\times\Sigma^i \rightarrow \mathbb{B}$ such that $c^i(T_1, b^i_j)$ counts the number of letter $b^i_j$ in coding tree $T_1$.

The tree kernel on the two trees ($T_1$ and $T_2$) with height $k$ after the root nodes are assigned labels is defined as:
\begin{equation}
WL\text{-}CT(T_1,T_2)=<\varphi_{CT}(T_1),\varphi_{CT}(T_2)>,
\end{equation}
where
\begin{align}
\varphi_{CT}(T_1)=&(c^0(T_1, b^0_1),\dots,c^0(T_1, b^0_{\left|\Sigma^0\right|}), \dots, \nonumber\\
& \,\,c^k(T_1, b^k_1),\dots,c^k(T_1, b^k_{\left|\Sigma^k\right|})) \nonumber
\end{align}
and
\begin{align}
\varphi_{CT}(T_2)=&(c^0(T_2, b^0_1),\dots,c^0(T_2, b^0_{\left|\Sigma^0\right|}),\dots, \nonumber \\
&\,\,c^k(T_2, b^k_1),\dots,c^k(T_2, b^k_{\left|\Sigma^k\right|})). \nonumber
\end{align}
\end{mydef}
Following the label counting process in the WL subtree kernel, our tree kernel is also designed to count the number of common labels in two coding trees.

\begin{mytheo}
\label{theo:treeKernelComplexity}
The WL-CT kernel on two coding trees $T_1$ and $T_2$ with the same height $k$ can be computed in time $O(n)$, which is much simpler than the WL subtree kernel ($O(hm)$) with $h$ iterations on $m$ edges \cite{shervashidze2011weisfeiler} and is the method in graph classification with the lowest time complexity to the best of our knowledge \cite{wu2020comprehensive}.
\end{mytheo}

\begin{myproof}
Given a graph $G$, the runtime of the WL subtree kernel is $O(hm)$ for $h$ iteration, as there are $O(m)$ elements in the multisets of a graph in each WL test iteration.
Correspondingly, given a coding tree $T$ for $G$, the time complexity of the WL-CT kernel is also the total number of elements in the multiset of the coding tree due to the similar label propagation scheme. In addition, the number of elements in the multiset is determined by the number of times that label propagation occurs, which is the edge number $|E|=m$. 
Thus, the time complexity of the WL-CT kernel is determined by the number of edges in $T$. As shown in Algorithm \ref{code:coding_tree}, the coding tree $T$ with the most edges is the full-height binary coding tree $T_B$, i.e., $O(T)\leq O(T_B)$. The complexity on the full-height binary coding tree is calculated as:
\begin{align}
\nonumber
O(T_B) & = O(|E_{T_B}|) \\
\nonumber & = O(|V_{T_B}|-1) \\
\nonumber & = O(|V_{T_B}^0| + |V_{T_B}^1|, \dots, +|V_{T_B}^{h_{max}}| - 1) \\
\nonumber & < O(2|V_{T_B}^0|) \\
\nonumber & = O(2|V|) \\
\nonumber & = O(n). 
\end{align}
\end{myproof}

\subsection{A Convolutional Network for Graph Classification}
Based on our tree kernel, we develop a novel graph convolutional network, \textit{HRN}, that generalizes the hierarchical reporting scheme to update the hidden features of nonleaf nodes for graph classification. HRN uses the hierarchical structure of coding tree and leaf node features $X_v$ to learn the representation vector of the entire coding tree $r_T$. HRN follows the proposed hierarchical reporting scheme, where the representation of a nonleaf node is updated by aggregating the hidden features of its children. Formally, the $i$-th layer of HRN is
\begin{equation}
\label{eq:etl_aggre}
r_v^i = \text{MLP}^i\left(\sum\nolimits_{u\in\mathcal{C}(v)}r_u^{(i-1)}\right),
\end{equation}
where $r^i_v$ is the feature vector of node $v$ with height $i$ in the coding tree, $r^0_v = x_v$, and $\mathcal{C}(v)$ is a set of child nodes of $v$.
As shown in Equation \ref{eq:etl_aggre}, we employ summation and multilayer perceptrons (MLPs) to perform hierarchical reporting in HRN. Sum aggregators have been theoretically proven to be injective over multisets and more powerful than mean and max aggregators \cite{xu2019powerful}. MLPs are capable of representing the compositions of functions because of the universal approximation theorem \cite{hornik1989multilayer,hornik1991approximation}.

For graph classification, the root node representation $r^k_v$ can be naively employed as the representation of the entire coding tree $r_T$. However, as discussed in \cite{xu2019powerful}, better results could be obtained from features in earlier iterations. To cover all hierarchical information, we employ hidden features from each height/iteration of the model. This is achieved by an architecture similar to the GIN \cite{xu2019powerful}, where we model the entire coding tree with layer representations concatenated across all heights/layers of the HRN structure:
\begin{align}
\label{eq:eml_readout}
r_T = \text{CONCAT}(&\text{LAYERPOOL}(\{r_v^i | v\in V_T^i\}) \nonumber \\ 
&| i=0,1,\dots,k),
\end{align}
where $r_v^i$ is the feature vector of node $v$ with height $i$ in coding tree $T$ and $k$ is the height of $T$. In HRN, LAYERPOOL in Equation \ref{eq:eml_readout} can be replaced with the summation or averaging of all node vectors within the same iterations, which generalizes our tree kernel.

\section{Experiments}
We validate the feasibility of performance improving while simplifying model complexity by comparing the experimental results of the WL-CT kernel and HRN with those of the most popular kernel-based methods and GNNs on graph classification tasks \footnote{The code of the WL-CT kernel and HRN can be found at \url{https://github.com/Wu-Junran/HierarchicalReporting}.}.

\paragraph{Datasets.} We conduct graph classification on five benchmarks: three social network datasets (IMDB-BINARY, IMDB-MULTI, and COLLAB) and two bioinformatics datasets (MUTAG and PTC) \cite{morris2020tudataset} \footnote{Considering the limitations of coding trees on disconnected graphs, we only adopt datasets that do not contain such graphs.}. The data of the bioinformatic datasets and social network datasets differ in that the nodes in bioinformatics graphs have categorical labels that do not exist in social networks. Thus, the initial node labels for the tree kernel are organized as follows: the node degrees are taken as node labels for social networks, whereas the node degrees and node categorical labels are taken according to each bioinformatic dataset. Correspondingly, the initial node features of the HRN inputs are set to one-hot encodings of the node degrees for social networks and a combination of the one-hot encodings of the degrees and categorical labels for bioinformatic graphs. Table \ref{tab:test_acc} summarizes the characteristics of the five employed datasets.

\paragraph{Configurations.} Following \cite{xu2019powerful}, 10-fold cross-validation is conducted, and we present the average accuracies achieved to validate the performance of our methods in graph classification \footnote{Currently, there have been three kinds of configurations for experimental results; we elaborate on them in Appendix \url{https://github.com/Wu-Junran/HierarchicalReporting/blob/master/appendix.pdf}.}
Regarding the configuration of our tree kernel, we adopt the $C$-support vector machine ($C$-SVM) \cite{chang2011libsvm} as the classifier and tune the hyperparameter $C$ of the SVM and the height of the coding tree $\in[2, 3, 4, 5]$. We implement the classification program with an SVM from Scikit-learn, where we set another hyperparameter $\gamma$ as \textit{auto} for IMDB-BINARY and IMDB-MULTI and as \textit{scale} for COLLAB, MUTAG and PTC, and we set the other hyperparameters as their default values.

For configuration of HRN, the number of HRN iterations is consistent with the height of the associated coding trees, which is also $\in [2, 3, 4, 5]$. All MLPs have 2 layers, as in the setting of the GIN. For each layer, batch normalization is applied to prevent overfitting. We utilize the Adam optimizer and set its initial learning rate to 0.01. For a better fit, the learning rate decays by half every 50 epochs. Other tuned hyperparameters for HRN include the number of hidden dimensions $\in \{16, 32, 64\}$, the minibatch size $\in\{32, 128\}$, and the dropout ratio $\in\{0, 0.5\}$ after LAYERPOOL. The number of epochs for each dataset is selected based on the best accuracy within cross-validation results.
We apply the same layer-level pooling approach (LAYERPOOL in Eq. \ref{eq:eml_readout}) for HRN; specifically, sum pooling is conducted on the bioinformatics datasets, and average pooling is conducted on the social datasets to ensure a direct comparison with GIN-0.

\paragraph{Baselines.} It is worth noting that recent research efforts have been devoted to the adoption of attention or graph pooling mechanisms, which also sheds light on the path of our future work. However, little attention has been paid to the core part of graph classification models. Therefore, in this paper, we employ models without an attention mechanism or well-designed graph pooling for a fair comparison. We compare our WL-CT kernel and HRN configured above with several state-of-the-art baselines for graph classification: (1) kernel-based methods, i.e., the WL subtree kernel \cite{shervashidze2011weisfeiler} and AWE \cite{ivanov2018anonymous}; (2) state-of-the-art deep learning methods, i.e., DCNN \cite{atwood2016diffusion}, PATCHY-SAN \cite{niepert2016learning}, DGCNN \cite{zhang2018end}, GIN-0 \cite{xu2019powerful} and LP-GNN \cite{tiezzi2021deep}. The accuracies of the WL subtree kernel are derived from \cite{xu2019powerful}. For AWE and the deep learning baselines, we utilize the accuracies contained in their original papers. 

\begin{table}[!b]
\centering
\resizebox{0.48\textwidth}{!}{
\begin{tabular}{lccccc}
\hline
Dataset & IMDB-B & IMDB-M & COLLAB & MUTAG & PTC \\
\# Graphs & 1000 & 1500 & 5000 & 188 & 344 \\
\# Classes & 2 & 3 & 3 & 2 & 2 \\
Avg. \# Nodes & 19.8 & 13.0 & 74.5 & 17.9 & 25.5 \\ \hline
\multicolumn{6}{c}{Kernel-based methods} \\ \hline
WL & 73.8$\pm$3.9 & 50.9$\pm$3.8 & 78.9$\pm$1.9 & 90.4$\pm$5.7 & 59.9$\pm$4.3 \\
AWE & 74.5$\pm$5.9 & 51.5$\pm$3.6 & 73.9$\pm$1.9 & 87.9$\pm$9.8 &  \\
\textbf{WL-CT} & \textbf{74.7$\pm$3.5} & \textbf{52.4$\pm$4.5} & \textbf{81.5$\pm$1.2} & \textbf{89.5$\pm$6.1} & \textbf{63.7$\pm$4.7} \\ \hline
\multicolumn{6}{c}{Deep learning methods} \\ \hline
DCNN & 49.1 & 33.5 & 52.1 & 67.0 & 56.6 \\
PATCHY-SAN & 71.0$\pm$2.2 & 45.2$\pm$2.8 & 72.6$\pm$2.2 & \textbf{92.6$\pm$4.2} & 60.0$\pm$4.8 \\
DGCNN & 70.0 & 47.8 & 73.7 & 85.8 & 58.6 \\
GIN-0 & 75.1$\pm$5.1 & 52.3$\pm$2.8 & 80.2$\pm$1.9 & 89.4$\pm$5.6 & 64.6$\pm$7.0 \\
LP-GNN-Single & 65.3$\pm$4.7 & 46.6$\pm$3.7 &  & 90.5$\pm$7.0 & 64.4$\pm$5.9 \\
LP-GNN-Multi & 76.2$\pm$3.2 & 51.1$\pm$2.1 &  & 92.2$\pm$5.6 & \textbf{67.9$\pm$7.2} \\
\textbf{HRN} & \textbf{77.5$\pm$4.3} & \textbf{52.8$\pm$2.7} & \textbf{81.8$\pm$1.2} & \textbf{90.4$\pm$8.9} & \textbf{65.7$\pm$6.4} \\ \hline
\end{tabular}}
\caption{\textbf{Classification accuracies on five benchmarks (\%).} The best results are highlighted in boldface. On datasets where WL-CT and HRN are not strictly the highest-scoring models among the baselines, our methods still achieve competitive results; thus, their accuracies are still highlighted in boldface. For baselines, we highlight those that are significantly higher than those of all other methods.}
\label{tab:test_acc}
\end{table}

\section{Results}
\label{sec:results}
The results of validating the WL-CT kernel and HRN on graph classification tasks are presented in Table \ref{tab:test_acc}. Our methods are shown in boldface. In the panel of kernel-based methods, we can observe that the accuracies of the WL-CT kernel exceed those of other kernel-based methods on four out of five benchmarks. For the only failed dataset, MUTAG, the WL-CT kernel still achieves very competitive performance. Notably, the WL-CT kernel even outperforms the state-of-the-art deep learning method (i.e., GIN-0) on IMDB-MULTI, COLLAB and MUTAG, which implies that superior performance can sometimes be obtained through a traditional machine learning methods rather than a deep learning model.

In the lower panel containing the deep learning methods, we can observe that the results of HRN are naturally superior to the accuracies of the WL-CT kernel, which further confirms the excellent performance of neural networks. In addition, HRN also yields the best performance on three out of five datasets, while competitive performance can still be observed on the other two datasets. In particular, the best performance of HRN is achieved on three social network datasets, which means social networks are more redundant than bioinformatic networks and that the key structure decoded from social networks is more conducive to model learning. In the light of the runtime complexity and experimental accuracies presented above, we can conclude that there is a way to boost the performance with simplified learning algorithms.

\begin{table}[!b]
\centering
\resizebox{0.48\textwidth}{!}{
\begin{tabular}{lcccccc}
\hline
LAYERPOOL & $\mathbb{H}$ & IMDB-B & IMDB-M & COLLAB & MUTAG & PTC \\ \hline
\multirow{4}{*}{ROOT} & 2 & 75.7$\pm$5.3 & 52.5$\pm$3.0 & \textbf{82.2$\pm$1.4} & 89.5$\pm$8.4 & 62.4$\pm$9.9 \\
 & 3 & 75.3$\pm$5.9 & 52.5$\pm$2.5 & 81.6$\pm$1.3 & 88.4$\pm$6.0 & 59.8$\pm$5.3 \\
 & 4 & 75.3$\pm$5.6 & 52.2$\pm$3.6 & 80.5$\pm$1.3 & 88.8$\pm$8.6 & 63.6$\pm$5.8 \\
 & 5 & 75.2$\pm$5.3 & 51.5$\pm$3.8 & 79.9$\pm$1.1 & 89.5$\pm$6.5 & 62.6$\pm$7.9 \\ \hline
\multirow{4}{*}{SUM} & 2 & 76.7$\pm$3.3 & 52.3$\pm$3.1 & 82.1$\pm$0.9 & 89.3$\pm$6.4 & 64.0$\pm$6.3 \\
 & 3 & \textbf{77.6$\pm$4.1} & 52.3$\pm$2.3 & 81.0$\pm$0.8 & 90.4$\pm$8.9 & 62.2$\pm$5.6 \\
 & 4 & 76.6$\pm$3.5 & 52.7$\pm$1.4 & 80.0$\pm$1.4 & 90.0$\pm$8.6 & \textbf{65.7$\pm$6.4} \\
 & 5 & 75.0$\pm$2.5 & 51.7$\pm$2.1 & 79.2$\pm$0.9 & 87.2$\pm$4.4 & 63.7$\pm$7.1 \\ \hline
\multirow{4}{*}{AVERAGE} & 2 & 76.9$\pm$2.9 & 52.6$\pm$3.3 & 81.8$\pm$1.2 & \textbf{91.4$\pm$6.1} & 61.6$\pm$8.9 \\
 & 3 & 77.5$\pm$4.3 & \textbf{52.8$\pm$2.7} & 81.2$\pm$1.3 & 88.8$\pm$5.6 & 61.0$\pm$7.9 \\
 & 4 & 76.4$\pm$3.0 & 52.6$\pm$2.1 & 80.5$\pm$1.6 & 87.2$\pm$7.2 & 64.4$\pm$9.9 \\
 & 5 & 75.5$\pm$2.5 & 52.2$\pm$2.1 & 79.6$\pm$0.9 & 87.8$\pm$5.8 & 63.1$\pm$7.4 \\ \hline
HRN-SOTA & \multicolumn{1}{l}{} & \multicolumn{1}{l}{77.6$\pm$4.1} & \multicolumn{1}{l}{52.8$\pm$2.7} & \multicolumn{1}{l}{82.2$\pm$1.4} & \multicolumn{1}{l}{91.4$\pm$6.1} & \multicolumn{1}{l}{65.7$\pm$6.4} \\ \hline
\end{tabular}}
\caption{\textbf{Graph classification accuracies of HRN with different LAYERPOOL approaches (\%).} The best results of each benchmark are highlighted in boldface. $\mathbb{H}$ denotes the coding tree height.}
\label{tab:hrn-lp}
\end{table}

\paragraph{Variants of HRN with different LAYERPOOL approaches.} In the experimental setup, we employ sum pooling for the bioinformatics datasets and average pooling for social network datasets to ensure a fair comparison with GIN-0. However, as described in the paper of GIN, sum pooling on bioinformatics datasets and average pooling on social datasets are adopted due to their better test performance \cite{xu2019powerful}. To explore the upper limits of HRN, we adopt various basic pooling approaches for the five benchmarks. In particular, in addition to the adopted sum and average pooling, we also select the hidden feature of the root node as the representation of the entire coding tree, denoted ROOT. The classification accuracies of HRN with three kinds of LAYERPOOL approaches for coding trees under different heights are shown in Table \ref{tab:hrn-lp}. With the exception of PTC and IMDB-MULTI, HRN does not achieve the optimal performance for the other three benchmarks with the LAYERPOOL selection of GIN-0. Specifically, HRN obtains the highest accuracies of 77.6\% for IMDB-BINARY with sum pooling, 82.2\% for COLLAB with ROOT, and 91.4\% for MUTAG with average pooling. In particular, the best result of HRN on COLLAB is realized with the root node representation rather than the sum or average pooling designed with skip connections for better performance. The reason for this phenomenon may be that the model overfits the detailed information in the shallow layers, which leads to misjudgment of the global property of graphs, especially considering that the average number of nodes in COLLAB is the largest (see Table \ref{tab:test_acc}); specifically, more detailed information is encountered. In summary, these results of HRN with different pooling methods indicate the great potential of HRN in graph representation learning.

\paragraph{The guidance of structural entropy.} Despite the state-of-the-art performance of our methods, we further evaluate the effectiveness of our $k$-dimensional coding tree algorithm. In particular, we do not form coding trees with the structural entropy minimization but use only a random coding tree, i.e., a randomly balanced binary tree (RBBT) with a height of two. For each graph within datasets, we produce the corresponding RBBT with graph nodes as leaves and employ our designed WL-CT kernel and HRN to perform graph classification (i.e., WL-CT-RBBT and HRN-RBBT). To ensure a fair comparison, we also fix the height of the coding tree to two (i.e., WL-CT-H2 and HRN-H2).
The results are shown in Table \ref{tab:RBBT-CT}. As can be seen, the structural entropy-guided coding tree surpasses the random coding tree not only on kernel-based methods but also on deep learning methods. The RBBT obtains higher accuracy only on the IMDB-BINARY dataset with kernel-based methods. 

\begin{table}[!t]
\centering
\resizebox{0.48\textwidth}{!}{
\begin{tabular}{lccccc}
\hline
Dataset & IMDB-B & IMDB-M & COLLAB & MUTAG & PTC \\ \hline
WL-CT-RBBT & 74.8$\pm$4.6 & 51.3$\pm$3.0 & 80.6$\pm$1.5 & 88.9$\pm$6.0 & 60.8$\pm$9.9 \\
WL-CT-H2 & 74.1$\pm$4.7 & 52.4$\pm$4.5 & 80.8$\pm$1.4 & 89.5$\pm$6.1 & 61.5$\pm$7.3 \\ \hline
HRN-RBBT & 73.0$\pm$2.4 & 52.0$\pm$2.4 & 78.7$\pm$1.0 & 87.2$\pm$6.9 & 63.1$\pm$4.2 \\
HRN-H2 & 76.9$\pm$2.9 & 52.6$\pm$3.3 & 81.8$\pm$1.2 & 89.3$\pm$6.4 & 64.0$\pm$6.3 \\ \hline
\end{tabular}}
\caption{Classification accuracies of the RBBT and coding tree with a height of two on five benchmarks (\%).}
\label{tab:RBBT-CT}
\end{table}

\begin{table}[!b]
\centering
\resizebox{0.48\textwidth}{!}{
\begin{tabular}{ll|cccccc}
\hline
Graph Kernel & Data Type & IMDB-BINARY & IMDB-MULTI & COLLAB & MUTAG & PTC & Higher \\ \hline
\multirow{2}{*}{Core Framework} & Graph & 73.7$\pm$5.3 & 53.2$\pm$5.0 & 81.0$\pm$1.8 & 86.2$\pm$4.8 & 63.1$\pm$4.1 & \textbf{3/5} \\
 & Coding Tree & 74.6$\pm$5.4 & 52.5$\pm$4.2 & 80.2$\pm$1.6 & 89.9$\pm$6.0 & 60.4$\pm$7.5 & 2/5 \\ \hline
\multirow{2}{*}{Neighborhood Hash} & Graph & 68.9$\pm$4.6 & 49.1$\pm$4.5 & 71.3$\pm$1.6 & 87.3$\pm$5.8 & 62.8$\pm$7.3 & 1/5 \\
 & Coding Tree & 76.4$\pm$4.5 & 51.8$\pm$3.8 & 81.2$\pm$1.2 & 87.8$\pm$7.5 & 61.3$\pm$9.2 & \textbf{4/5} \\ \hline
\multirow{2}{*}{ODD-STh} & Graph & 72.8$\pm$3.4 & 50.9$\pm$4.4 & 79.9$\pm$1.9 & 83.4$\pm$9.1 & 63.4$\pm$8.5 & 1/5 \\
 & Coding Tree & 74.0$\pm$4.7 & 51.5$\pm$4.9 & 80.1$\pm$1.9 & 87.3$\pm$5.8 & 62.8$\pm$5.8 & \textbf{4/5} \\ \hline
\multirow{2}{*}{Propagation} & Graph & 73.1$\pm$4.8 & 51.5$\pm$4.0 & 79.1$\pm$1.4 & 84.1$\pm$6.6 & 60.8$\pm$5.5 & 2/5 \\
 & Coding Tree & 75.3$\pm$5.2 & 50.9$\pm$4.3 & 78.3$\pm$1.0 & 86.2$\pm$4.8 & 62.5$\pm$6.8 & \textbf{3/5} \\ \hline
\multirow{2}{*}{Pyramid Match} & Graph & 75.7$\pm$4.5 & 51.4$\pm$4.5 & 81.9$\pm$1.2 & 88.9$\pm$7.9 & 61.6$\pm$8.0 & 1/5 \\
 & Coding Tree & 74.5$\pm$4.9 & 51.8$\pm$4.8 & 82.2$\pm$1.7 & 89.9$\pm$7.7 & 65.1$\pm$8.5 & \textbf{4/5} \\ \hline
\multirow{2}{*}{Shortest Path} & Graph & 74.4$\pm$4.0 & 53.2$\pm$5.2 & 80.0$\pm$1.9 & 86.7$\pm$4.3 & 62.2$\pm$4.2 & 2/5 \\
 & Coding Tree & 74.6$\pm$5.4 & 52.5$\pm$4.2 & 80.6$\pm$1.8 & 89.9$\pm$6.0 & 60.4$\pm$7.5 & \textbf{3/5} \\ \hline
\end{tabular}%
}\caption{Classification accuracies of coding trees and original graphs within various graph kernels on 5 benchmarks (\%).}
\label{tab:ct_g_kernels}
\end{table}

\paragraph{The effectiveness of essential structure within graphs.} Following the label propagation scheme within the WL subtree kernel \cite{shervashidze2011weisfeiler}, we present a WL-CT kernel to take full advantage of the hierarchical structure of coding trees and validate its superior performance on five benchmarks. Here, we delve deeper into the effectiveness of the essential structure within graphs. Specifically, we employ those classical graph kernels, including the Core Framework kernel, Neighborhood Hash kernel, ODD-STh kernel, Propagation kernel, Pyramid Match kernel, and Shortest Path kernel, rather than the WL-CT kernel to perform classification. Note that the nonleaf nodes within coding trees are virtual nodes without initial labels; thus, we assign all nonleaf nodes in coding trees with the same label $0$. As shown in Table \ref{tab:ct_g_kernels}, graphs and coding trees have distinct advantages under different kernels, while coding trees can still achieve superior performance even with these kernels designed for graphs with the exception of the core framework kernel, which further confirms that the coding trees decoded by structural entropy minimization are key structures of given graphs.

\begin{figure}[!t]
  \centering
  \includegraphics[width=0.48\textwidth]{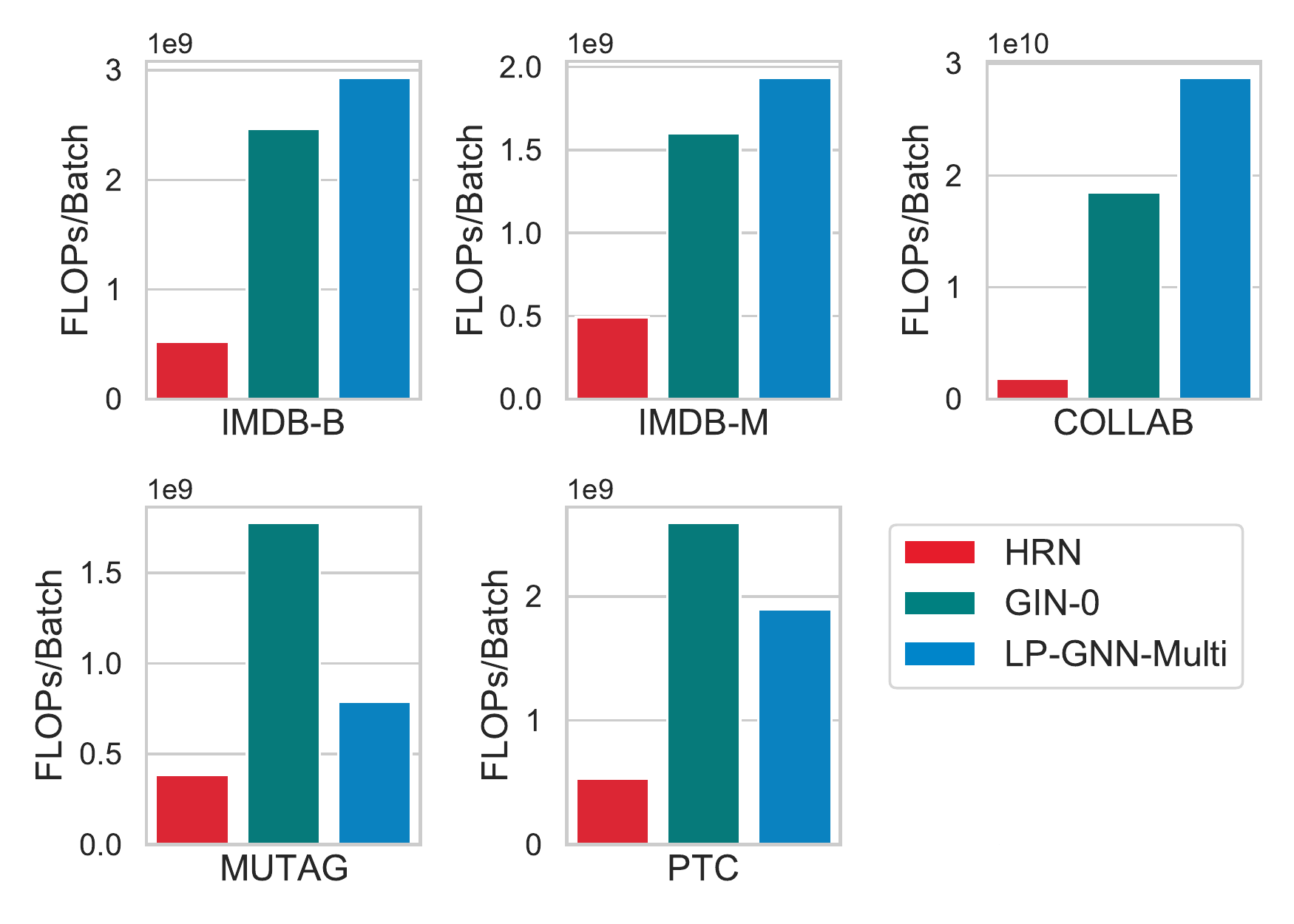}   
  \caption{Comparison regarding the volume of computations.} 
  \label{fig:log_flops}
\end{figure}

\paragraph{Computational efficiency.} In addition to the lower runtime complexity of our HRN than that of state-of-the-art GNNs ($O(n) < O(hm)$), we further compare the volumes of computations, i.e., the numbers of floating-point operations per second (FLOPs), required by HRN, GIN-0 and LP-GNN-Multi under the same parameter settings on all datasets. Specifically, we fix the number of iterations to four (the five GNN layers of GIN-0 include the input layer), the number of hidden units to 32, the batch size to 128 and the final dropout ratio to 0. The results are shown in Fig.~\ref{fig:log_flops}. One can see that the volume of computations required by HRN is consistently smaller than that of GIN-0 and LP-GNN-Multi. More concretely, HRN needs only 20\% (25\%) of the volume of computations needed by the GIN-0 (LP-GNN-Multi) on average.

\section{Conclusion}
In this paper, we investigate the feasibility of improving graph classification performance while simplifying the learning process. In particular, through structural entropy minimizing, we decode the essential structure of given graphs to a simpler data form, i.e., coding trees. With this optimized new structure, we improve upon the graph classification performance by simplifying the corresponding kernel method and deep learning method. In summary, our proposed WL-CT kernel and HRN are not only theoretically and experimentally much more efficient but also capable of achieving state-of-the-art performance on most benchmarks.
In addition to the excellent graph classification performance, out methods are not fit for another important task in the graph realm, i.e., node classification. Hence, determining how to improve the performance of node classification with a simpler learning model may be another underlying direction for future work.

\section*{Acknowledgments}
This research was supported by NSFC (Grant No. 61932002).

\newpage
\bibliographystyle{named}
\bibliography{ijcai22}

\end{document}